\documentclass[11pt]{article}
\usepackage{nodalida2021}
\usepackage{times}
\usepackage{url}
\usepackage{latexsym}
\usepackage{hyperref}

\usepackage{verbatim}
\usepackage{latexsym}
\usepackage{graphicx}

\usepackage{microtype}

\aclfinalcopy 

\title{Should we Stop Training More Monolingual Models, \\
and Simply Use Machine Translation Instead?}

\author{Tim Isbister \\
  Peltarion\\
  {\small {\tt tim.isbister@peltarion.com}} \\\And
  Fredrik Carlsson \\
  RISE \\
  {\small {\tt fredrik.carlsson@ri.se}} \\\And
  Magnus Sahlgren \\
  RISE \\
  {\small {\tt magnus.sahlgren@ri.se}} \\}

\date{}

\begin{document}
\maketitle

\begin{abstract}
Most work in NLP makes the assumption that it is desirable to develop solutions in the native language in question. There is consequently a strong trend towards building native language models even for low-resource languages. This paper questions this development, and explores the idea of simply translating the data into English, thereby enabling the use of pretrained, and large-scale, English language models.
We demonstrate empirically that a large English language model coupled with modern machine translation outperforms native language models in most Scandinavian languages. The exception to this is Finnish, which we assume is due to inferior translation quality.
Our results suggest that machine translation is a mature technology, which raises a serious counter-argument for training native language models for low-resource languages.
This paper therefore strives to make a provocative but important point. 
As English language models are improving at an unprecedented pace, which in turn improves machine translation, it is from an empirical and environmental stand-point more effective to translate data from low-resource languages into English, than to build language models for such languages.
\end{abstract}

\section{Introduction}
Although the Transformer architecture for deep learning was only recently introduced \cite{NIPS2017_3f5ee243}, it has had a profound impact on the development in Natural Language Processing (NLP) during the last couple of years. Starting with the seminal BERT model \cite{devlin-etal-2019-bert}, we have witnessed an unprecedented development of new model variations \cite{NEURIPS2019_dc6a7e65,Clark2020ELECTRA:,2020t5,radford2019language,brown2020language} with new State Of The Art (SOTA) results being produced in all types of NLP benchmarks \cite{wang-etal-2018-glue,NEURIPS2019_superglue,nie-etal-2020-adversarial}.

The leading models are large both with respect to the number of parameters and the size of the training data used to build the model; this correlation between size and performance has been demonstrated by \citet{kaplan2020scaling}. The ongoing scale race has culminated in the 175-billion parameter model GPT-3, which was trained on some 45TB of data summing to around 500 billion tokens \cite{brown2020language}.\footnote{The currently largest English model contains $1.6$ trillion parameters \citep{SwitchTransformer}.} Turning to the Scandinavian languages, there are no such truly large-scale models available. At the time of writing, there are around 300 Scandinavian models available in the Hugging Face Transformers model repository.\footnote{\url{huggingface.co/models}} Most of these are translation models, but there is already a significant number of monolingual models available in the Scandinavian languages.\footnote{At the time of submission, there are 17 monolingual Swedish models available.} 

However, none of these Scandinavian language models are even close to the currently leading English models in parameter size or training data used. As such, we can expect that their relative performance in comparison with the leading English models is significantly worse. Furthermore, we can expect that the number of monolingual Scandinavian models will continue to grow at an exponential pace during the near future. The question is: do we need all these models? Or even: do we need {\em any} of these models? Can't we simply translate our data and tasks to English and use some suitable English SOTA model to solve the problem? This paper provides an empirical study of this idea.

\begin{table*}[t]
	\begin{center}
		\begin{tabular}{|l|rrrr|}
			\hline
			{\bf Language}  & {\bf Vocab size} & {\bf Lexical richness} & {\bf Avg. word length} & {\bf Avg. sentence length} \\
			\hline
			Swedish   & 31,478          & 0.07                & 4.39               & 14.75                   \\
			Norwegian & 26,168          & 0.06                & 4.21               & 14.10                   \\
			Danish    & 42,358          & 0.06                & 4.17               & 19.55                   \\
			Finnish   & 34,729          & 0.14                & 5.84               & 10.69                   \\
			\hline
			English   & 27,610          & 0.04                & 3.99               & 16.87                   \\
			\hline
		\end{tabular}
		\caption{The vocabulary size, Lexical richness, average word length and average sentence length for the Trustpilot sentiment data of each language.}
		\label{tab:dataset_1}
	\end{center}
\end{table*}

\section{Related work}

There is already a large, and rapidly growing, literature on the use of multilingual models \cite{conneau-etal-2020-unsupervised,xue2020mt5}, and on the possibility to achieve cross-lingual transfer in multilingual language models \cite{ruder2019survey,artetxe-etal-2020-cross,lauscher-etal-2020-zero,conneau-etal-2020-emerging,K2020Cross-Lingual,nooralahzadeh2020meta}. From this literature, we know among other things that multilingual models tend to be competitive in comparison with monolingual ones, and that especially languages with smaller amounts of training data available can benefit significantly from transfer effects from related languages with more training data available. This line of study focuses on the possibility to transfer {\em models} to a new language, and thereby facilitating the application of the model to data in the original language. 

By contrast, our interest is to transfer the {\em data} to another language, thereby enabling the use of SOTA models to solve whatever task we are interested in. We are only aware of one previous study in this direction: \citet{IsMTRipe} performs cross-lingual machine translation using outdated methods, resulting in the claim that even if perfect translation would be possible, we will still see degradation of performance. In this paper, we use modern machine translation methods, and demonstrate empirically that no degradation of performance is observable when using large SOTA models.

\begin{table*}[!ht]
	\centering
	\begin{tabular}{|l|r|r|}
		\hline
		{\bf Model name in Hugging Face}                       & {\bf Language} & {\bf Data size} \\
		\hline
		\texttt{KB/bert-base-swedish-cased}                       & sv       & 3B tokens \\
		\texttt{TurkuNLP/bert-base-finnish-cased-v1}              & fi       & 3B tokens \\
		\texttt{ltgoslo/norbert}                                  & no       & 2B tokens \\
		\texttt{DJSammy/bert-base-danish-uncased\_BotXO,ai}       & da       & 1.6B tokens \\
		\hline
		\texttt{bert-base-cased}                                  & en       & 3.3B tokens \\
		\texttt{bert-base-cased-large}                            & en       & 3.3B tokens \\
\hline
		\texttt{xlm-roberta-large}                                & multi    & 295B tokens \\
		\hline
	\end{tabular}
	\caption{Models used in the experiments and the size of their corresponding training data. 'B' is short for billion.}
	\label{tab:models}
\end{table*}

\begin{table*}[!htbp]
	\begin{center}
		\begin{tabular}{|l|r|r|r|r|r|r|}
			\hline
			\textbf{Model} & {\bf sv}                 & {\bf no}                  & {\bf da} & {\bf fi} & {\bf en}             \\
			\hline
			BERT-sv        & \underline{96.76}     & 89.32               & 90.68                & 83.40                & 86.76          \\
			BERT-no        & 90.40                & \underline{95.00}             & 92.52                    & 83.16                   & 78.52              \\

			BERT-da        & 86.24              & 89.16               & \underline{94.72}                & 80.16               & 85.28              \\
			BERT-fi        & 90.24              & 86.36               & 87.72                & \underline{\textbf{95.72}}    & 84.32              \\
			\hline
			BERT-en        & 85.72              & 87.60                & 87.72                & 84.16               & 96.08     \\

			BERT-en-Large  & 91.16              & 91.88              & 92.40                  & 89.56               & \underline{{\bf 97.00}}     \\
			\hline
            \hline
			\multicolumn{6}{|c|}{\textbf{Translated Into English}} \\
			\hline
			BERT-sv        & 88.24          & 87.80      & 89.68      & 83.60        & -          \\
			BERT-no        & 88.40          & 86.80      & 88.44      & 80.72        & -          \\
			BERT-da        & 88.24          & 84.20      & 89.12      & 83.32        & -          \\
			BERT-fi        & 90.04          & 90.08     & 89.36       & 86.04        & -          \\
			\hline
			BERT-en        & 95.76          & 95.48     & 95.96       & 92.96        & -          \\
			BERT-en-Large  & {\bf 97.16}    & {\bf 96.56}    & {\bf 97.48}      & 94.84        & -          \\
			\hline
		\end{tabular}
		\caption{Accuracy for monolingual models for the native sentiment data (upper part) and machine translated data (lower part). Underlined results are the best results per language in using the native data, while boldface marks the best results considering both native and machine translated data.}
		\label{tab:original}
	\end{center}
\end{table*}

\begin{table*}[!htbp]
\begin{center}
\begin{tabular}{|l|r|r|r|r|r|r|}
\hline
\textbf{Model} & {\bf sv} & {\bf no} & {\bf da} & {\bf fi} & {\bf en} \\
\hline
XLM-R-large & \textbf{97.48} & \textbf{97.16} & 97.68 & 95.60 & \textbf{97.76} \\
\hline
\multicolumn{6}{|c|}{\textbf{Translated Into English}} \\
\hline
XLM-R-large & 97.04 & 96.84 & \textbf{98.24} & 95.48 & - \\
\hline
\end{tabular}
\caption{Accuracy on the various sentiment datasets using XLM-R-Large}
\label{tab:xlm-r}
\end{center}
\end{table*}

\section{Data}
In order to be able to use comparable data in the languages under consideration (Swedish, Danish, Norwegian, and Finnish), we contribute a Scandinavian sentiment corpus (ScandiSent),\footnote{https://github.com/timpal0l/ScandiSent}
consisting of data downloaded from \url{trustpilot.com}. For each language, the corresponding subdomain was used to gather reviews with an associated text. This data covers a wide range of topics and are divided into 22 different categories, such as electronics, sports, travel, food, health etc. The reviews are evenly distributed among all categories for each language.

All reviews have a corresponding rating in the range $1-5$.
The review ratings were polarised into binary labels, and the reviews which received neutral rating were discarded. Ratings with 4 or 5 thus corresponds to a positive label, and 1 or 2 correspond to a negative label.

To further improve the quality of the data, we apply fastText's language identification model \cite{joulin2016bag} to filter out any reviews containing incorrect language. This results in a balanced set of 10,000 texts for each language, with 7,500 samples for training and 2,500 for testing. Table \ref{tab:dataset_1} summarizes statistics for the various datasets of each respective language.

\subsection{Translation}
For all the Nordic languages we generate a corresponding English dataset by direct Machine Translation, using the Neural Machine Translation (NMT) model provided by Google.\footnote{https://cloud.google.com/translate/docs/advanced/translating-text-v3}
To justifiably isolate the effects of modern day machine translation, we restrict the translation to be executed in prior to all experiments. This means that all translation is executed prior to any fine-tuning, and that the translation model is not updated during training.

\section{Models}

In order to fairly select a representative pre-trained model for each considered Scandinavian language, we opt for the most popular native model according to Hugging Face. For each considered language, this corresponds to a BERT-Base model, hence each language is represented by a Language Model of identical architecture. The difference between these models is therefore mainly in the quantity and type of texts used during training, in addition to potential differences in training hyperparameters.

We compare these Scandinavian models against the English BERT-Base and BERT-Large models by Google. English BERT-Base is thus  identical in architecture to the Scandinavian models, while BERT-Large is twice as deep and contains more than three times the amount of parameters as BERT-Base. Finally, we include XLM-R-Large, in order to compare with a model trained on significantly larger (and multilingual) training corpora. 

Table \ref{tab:models} lists both the Scandinavian and English models, together with the size of each models corresponding training corpus.

\section{Experiments}
\subsection{Setup}

We fine-tune and evaluate each model towards each of the different sentiment datasets, using the hyperparameters listed in Appendix \ref{tab:parameters}. From this we report the binary accuracy, with the results for the BERT models available in Table \ref{tab:original}, and the XLM-R results in Table \ref{tab:xlm-r}. 

\subsection{Monolingual Results}

The upper part of Table \ref{tab:original} shows the results using the original monolingual data.
From this we note a clear diagonal (marked by underline), where the native models perform best in their own respective language. Bert-Large significantly outperforms BERT-Base for all non-English datasets, and it also performs slightly better on the original English data.

Comparing these results with the amount of training data for each model (Table \ref{tab:dataset_1}), we see a correlation between performance and amount of pre-training data. The Swedish, Finnish and English models have been trained on the most amount of data, leading to slightly higher performance in their native languages. The Danish model which has been trained on the least amount of data, performs the worst on its own native language.

For the cross-lingual evaluation, BERT-Large clearly outperforms all other non-native models. The Swedish model reaches higher performance on Norwegian and Finnish compared to the other non-native Scandinavian models. However, the Norwegian model performs best of the non-native models on the Danish data. Finally, we observe an interesting anomaly in the results on the English data, where the Norwegian model performs considerably worse than the other Scandinavian models.

\subsection{Translation Results}

The results for the machine translated data, available as the lower part of Table \ref{tab:original}, show that BERT-Large outperforms all native models on their native data, with the exception of Finish. The English BERT-Base reaches higher performance on the machine translated data than the Norwegian and Danish models on their respective native data. The difference between English BERT-Base using the machine translated data, and the Swedish BERT using native data is about 1\% unit.


As expected, all Scandinavian models perform significantly worse on their respective machine translated data. We find no clear trend among the Scandinavian models when evaluated on translated data from other languages. But we note that the Danish model performs better on the machine translated Swedish data than on the original Swedish data, and the Finnish model also improves its performance on the other translated data sets (except for Swedish). All models (except, of course, the Finnish model) perform better on the machine translated Finnish data. 

Finally, \ref{tab:xlm-r} shows the results from XLM-R-Large, which has been trained on data several orders of magnitude larger than the other models. XLM-R-Large achieves top scores on the sentiment data for all languages except for Finnish. We note that XLM-R produces slightly better results on the native data for Swedish, Norwegian and Finnish, while the best result for Danish is produced on the machine translated data.


\section{Discussion \& Conclusion}

Our experiments demonstrate that it is possible to reach better performance in a sentiment analysis task by translating the data into English and using a large pre-trained English language model, compared to using data in the original language and a smaller native language model. Whether this result holds for other tasks as well remains to be shown, but we see no theoretical reasons for why it would not hold. We also find a strong correlation between the quantity of pre-training data and downstream performance. We note that XLM-R in particular performs well, which may be due to data size, and potentially the ability of the model to take advantage of transfer effects between languages.

An interesting exception in our results is the Finnish data, which is the only task for which the native model performs best, despite XLM-R reportedly having been trained on more Finnish data than the native Finnish BERT model \cite{conneau-etal-2020-unsupervised}. One hypothesis for this behavior can be that the alleged transfer effects in XLM-R hold primarily for typologically similar languages, and that the performance on typologically unique languages, such as Finnish, may actually be negatively affected by the transfer. The relatively bad performance of BERT-Large on the translated Finnish data is likely due to insufficient quality of the machine translation. 

The proposed approach is thus obviously dependent on the existence of a high-quality machine translation solution. The Scandinavian languages are typologically very similar both to each other and to English, which probably explains the good performance of the proposed approach even when using a generic translation API. For other languages, such as Finnish in our case, one would probably need to be more careful in selecting a suitable translation model. Whether the suggested methodology will be applicable to other language pairs thus depends on the quality of the translations and on the availability of large-scale language models in the target language.

Our results can be seen as evidence for the maturity of machine translation. Even using a generic translation API, we can leverage the existence of large-scale English language models to improve the performance in comparison with building a solution in the native language. This raises a serious counter-argument for the habitual practice in applied NLP to develop native solutions to practical problems. Hence, we conclude with the somewhat provocative claim that it might be unnecessary from an empirical standpoint to train models in languages where:

\begin{enumerate}
	\item there exists high-quality machine translation models to English,
	\item there does not exist as much training data to build a language model.
\end{enumerate}

In such cases, we may be better off relying on existing large-scale English models. This is a clear case for practical applications, where it would be beneficial to only host one large English model and translate all various incoming requests from different languages.

\bibliographystyle{acl_natbib}
\bibliography{nodalida2021}

\appendix
\section{Training Details}

\begin{table}[h]
\begin{center}
\begin{tabular}{|l|r|}
\hline
\textbf{Parameters}    & \textbf{Value}   \\
\hline
train\_epochs        & 2         \\
early\_stopping      & false     \\
optimizer            & AdamW     \\
learning\_rate       & 4e-5      \\
batch\_size          & 512       \\
max\_seq\_length	 & 128       \\
max\_grad\_norm      & 1.0       \\
\hline

\end{tabular}
\caption{Training hyperparameters for the sentiment classification experiments.}
\label{tab:parameters}
\end{center}
\end{table} 

\end{document}